\documentclass[
]{ceurart}

\usepackage{listings}
\begin{document}

\copyrightyear{2026}
\copyrightclause{Copyright for this paper by its authors.
  Use permitted under Creative Commons License Attribution 4.0
  International (CC BY 4.0).}

\conference{CLEF 2026 Working Notes, 21 -- 24 September 2026, Jena, Germany}

\title{Searching for Sound-Meaning Collisions: Graph-Based Affordance Retrieval and Multi-Evaluator Ranking for Pun Translation at CLEF 2026 JOKER Task 2}

\title[mode=sub]{Joker Task 2 at CLEF 2026}

\author[1]{Russell Taylor}[%
orcid=0009-0007-0702-2375,
email=rdtaylorjr@gatech.edu,
]
\cormark[1]
\address[1]{Georgia Institute of Technology, North Ave NW, Atlanta, GA 30332}

\author[1]{Adam Brikman}[%
orcid=0000-0000-0000-0000,
email=abrikman3@gatech.edu,
]

\author[1]{Prateek Awate}[%
orcid=0009-0002-1878-9700,
email=pawate3@gatech.edu,
]

\cortext[1]{Corresponding author.}

\begin{abstract}
Fifteen years ago, Low proposed that pun translators should stop searching for equivalent words and instead search for new points of contact between sound and meaning. In this paper, we investigate that idea computationally.

We model pun translation as a process of discovery, exploration, and selection. A retrieval system searches semantic and phonological neighborhoods for target-language affordances: sound--meaning bridges that may support new wordplay. Multiple language models then explore these opportunities by generating competing translations, while a multi-perspective generate-and-rank architecture selects among them.

Beyond system development, our primary contribution is an analysis of how retrieved affordances propagate through the translation process. We find that generators actively exploit retrieved opportunities, evaluators progressively concentrate around stronger sound--meaning bridges, and exact phonological collisions are selected at disproportionately high rates when available. At the same time, many puns still yield no usable affordances, suggesting that retrieval remains the central bottleneck in computational pun translation.

The resulting picture is remarkably close to the process envisioned by Low. Successful pun translation emerges not from preserving source-language words, but from discovering new places in the target language where sound and meaning collide.
\end{abstract}


\maketitle

\section{Introduction}

Pun translation is often described as impossible because the linguistic accident that creates a pun in one language rarely survives translation into another. Yet translators have long known that successful pun translation does not require preserving the original word. It requires finding another place in the target language where sound and meaning collide.

One particularly useful perspective was proposed by Low, who argued that translators should not search for equivalent words, but for new points of contact between sound and meaning. His squares, pentagons, and hexagons describe a progressively expanding search through semantic and phonological neighborhoods in pursuit of such opportunities. Although developed as a practical framework for professional human translators, the underlying process resembles a search procedure that lends itself naturally to computational implementation \cite{low2011translating}.

Our 2025 CLEF Joker Task 2 submissions, which achieved first and second place in human evaluation, were heavily influenced by this perspective. We combined semantic and phonetic retrieval with multi-agent evaluation and found that systems optimized for humor often outperformed approaches focused on literal translation \cite{taylor2025pun}. However, an important question remained unanswered: what role does retrieval actually play in successful pun translation?

In this work we treat pun translation as a process of discovery, exploration, and selection. Retrieval identifies semantic--phonetic affordances: target-language opportunities where sound and meaning may support new wordplay \cite{gibson1979ecological}. Generators explore these opportunities by producing multiple competing translations. Evaluators then select among them using multiple perspectives reflecting humor, wordplay quality, fluency, and preservation of the joke's underlying setup and humorous intent.

To support this investigation, we substantially expand the retrieval space through phrase-level semantic embeddings and learned phonetic embeddings, replace regeneration-based evaluation with a two-stage generate-and-rank architecture, and trace retrieved affordances throughout the entire translation pipeline. This allows us not only to improve performance, but also to study which sound--meaning bridges survive successive stages of generation and selection.

The results suggest that successful pun translation is neither direct translation nor unconstrained creativity. It is a search process. Retrieval uncovers hidden bridges between semantic domains, generators explore them, and evaluators progressively concentrate around the strongest collisions between sound and meaning.
\section{System Improvements}

\subsection{Preprocessing}

Consistent with Miller's proposed computational framework for pun translation, preprocessing aims to identify the source pun and its associated semantic domains before retrieval and candidate generation \cite{miller2019punsters}.

\subsubsection{Data Cleaning}

We revised our preprocessing pipeline to improve LLM comprehension of the source puns \cite{taylor2025pun}. Compared to our previous system, the cleaner normalizes Unicode punctuation, repairs malformed quotation and dialogue structures, preserves lexical hashtag cues that explicitly mark pun words, removes social-media artifacts (URLs, handles, hashtags, emojis, and reaction text), and incorporates approximately 100 dataset-specific repairs identified through automated validation. These changes were intended to improve downstream pun identification and interpretation rather than contribute directly to translation quality.

\subsubsection{Pun Word Identification and Translation}
We retained the pun word identification and translation stage from our previous system but simplified the output representation. The model identifies the pun word, pun type, and two semantic domain word lists associated with the wordplay. Unlike the previous system, we removed explicit context-word extraction and translate only the semantic fields required for downstream retrieval. We implemented the pipeline using gemini-3-flash and structured JSON outputs.

\subsection{Expanded Affordance Embedding Spaces}

A limitation of our previous system was that affordance discovery operated primarily at the word level. Many puns, however, rely on fixed expressions, collocations, and phrase-level reinterpretations that extend beyond individual lexical items. To address this limitation, we constructed a 370,450-entry French expression bank and indexed it for dense retrieval. 

\subsubsection{Expression-Level Semantic Embeddings}

The expression bank combines 51,589 French Wiktionary entries labeled as expressions, proverbs, or locutions \cite{wiktionary_latex}; 12,308 PARSEME-derived sentence contexts containing annotated multiword expressions \cite{savary2017parseme}; and 306,553 high-association OpenSubtitles collocations \cite{lison2018opensubtitles} mined using spaCy lemmatization, part-of-speech pattern filtering, frequency thresholds, and pointwise mutual information scoring \cite{spacy2}. After normalization and deduplication, 355,803 entries (96.1\%) consisted of multi-word units.

All entries were embedded using BAAI/bge-m3 \cite{chen-etal-2024-m3} with normalized embeddings and indexed using FAISS inner-product search \cite{johnson2017billion}. Unlike the previous system, which primarily searched for individual lexical affordances, the expression bank enables retrieval over multi-word expressions, collocations, idiomatic constructions, and longer semantic contexts. These additions expand the search space available to the retrieval system and provide phrase-level semantic anchors capable of supporting target-language wordplay.

\subsubsection{Learned Phonetic Embeddings}

Starting from a French lexicon containing orthographic forms and IPA transcriptions, we constructed a large-scale phonetic relevance graph containing 245,746 unique word--IPA pairs and 4.46 million labeled phonological relations \cite{sharma2021phonetic}. Positive relations included exact homophones, near-homophones, schwa-deletion variants, rhotic variants, nasalization variants, rhyme relations, consonant-skeleton relations, and vowel-skeleton relations; hard negatives were mined from phonologically similar forms that lacked meaningful phonological equivalence. The resulting dataset yielded 2.72 million positive training pairs and 355,007 held-out evaluation relations. Relation counts are summarized in Table~\ref{tab}.

\begin{table}[!htbp]
\centering
\caption{Positive relation types used for phonetic embedding.}
\label{tab}
\begin{tabular}{lr}
\toprule
Relation type & Count \\
\midrule
Strong rhyme & 1,518,818 \\
Vowel skeleton & 426,532 \\
Consonant skeleton & 421,482 \\
Rhotic variant & 73,139 \\
Near homophone (edit distance 2) & 66,379 \\
Nasalization variant & 57,013 \\
Exact homophone & 56,521 \\
Schwa deletion & 30,295 \\
Weak rhyme & 25,761 \\
Near homophone (edit distance 1) & 8,198 \\
Boundary deletion & 323 \\
\midrule
Total & 2,684,461 \\
\bottomrule
\end{tabular}
\end{table}

Unlike traditional phonetic retrieval systems that focus primarily on exact pronunciation similarity, the supervision signal intentionally emphasizes broader phonological neighborhoods, including rhyme, consonant-skeleton, and vowel-skeleton relations, to support retrieval of approximate sound correspondences useful for pun construction. These relations are used to fine-tune BAAI/bge-m3 \cite{chen-etal-2024-m3} on IPA sequences using SentenceTransformers and Multiple Negatives Ranking Loss \cite{reimers-2020-multilingual-sentence-bert}. Training pairs are balanced through capped per-relation sampling to prevent dominant relation classes from overwhelming the embedding geometry. Training used a batch size of 128 for one epoch. The resulting embedding space was trained to capture phonological relationships useful for pun construction rather than exact pronunciation equivalence alone. The trained model embeds IPA representations into a dense retrieval space indexed with FAISS \cite{johnson2017billion} and supports approximate phonetic retrieval through cosine-similarity nearest-neighbor search.

We evaluated the learned phonetic encoder on an intrinsic retrieval benchmark containing 6,321 query IPA forms and 244,948 indexed items. Performance was strongest for near-homophone and consonant-skeleton relations, achieving Recall@10 of 0.642 and 0.371 respectively, while hard-negative retrieval remained near zero. These results indicate that the embedding space effectively captures pun-relevant phonetic similarity beyond exact homophone matching.

\begin{table}[!htbp]
\centering
\caption{Intrinsic phonetic retrieval evaluation.}
\label{tab:phonetic_eval}
\begin{tabular}{lrr}
\toprule
Relation & R@10 & MRR \\
\midrule
All & 0.394 & 0.191 \\
Near homophone (edit 1) & 0.642 & 0.293 \\
Consonant skeleton & 0.371 & 0.172 \\
Strong rhyme & 0.027 & 0.013 \\
Hard negative & 0.000 & 0.000 \\
\bottomrule
\end{tabular}
\end{table}

\subsection{Graph-Based Affordance Retrieval}

Following Low's proposal of pun translation as a search for exploitable sound--meaning collisions, we model retrieval as exploration of semantic and phonetic neighborhoods connecting two semantic domains rather than direct translation between individual lexical items. We refer to retrieved phonetic-semantic pairs as affordances because they provide opportunities for target-language pun construction rather than direct translations. 

We view the retrieval space as a heterogeneous graph whose vertices are lexical items and expressions. Semantic retrieval induces semantic edges, while phonetic retrieval induces phonetic edges. Affordance discovery corresponds to identifying bridge structures that connect the semantic regions associated with the two source-pun domains through combinations of semantic and phonetic relations.

Let \(V\) denote the set of French lexical items and expressions. Semantic retrieval induces semantic neighborhoods \(N_S(\cdot)\), while phonetic retrieval induces phonetic neighborhoods \(N_P(\cdot)\). Let \(A\) and \(B\) denote the two translated lists of words representing the distinct semantic domains of the source pun word. These serve as the seed sets for two semantic regions.

\subsubsection{Semantic Expansion}

Each semantic region is expanded independently through dense retrieval over the French expression bank and bounded FastText lexical expansion \cite{grave2018learning}. The resulting semantic neighborhoods are

\[
A_S=N_S(A),
\qquad
B_S=N_S(B).
\]

All retrieved candidates must possess an attested IPA representation before participating in subsequent phonetic retrieval. Detached phonetic recovery is defined as

\[
A_P=N_P(A\cup A_S),
\qquad
B_P=N_P(B\cup B_S),
\]

where \(N_P\) denotes phonetic-neighborhood expansion in the learned IPA embedding space.

\subsubsection{Bucketized Bridge Mining}

We define phonetic bridge retrieval as

\[
\Phi(X,Y)
=
\{(x,y)\in X\times Y :
\operatorname{ipa}(x)\neq\emptyset,\;
\operatorname{ipa}(y)\neq\emptyset,\;
\cos(\mathbf{e}_x,\mathbf{e}_y)\ge\tau
\}
\]

where \(\mathbf{e}_x\) and \(\mathbf{e}_y\) are the learned phonetic embeddings of \(x\) and \(y\), respectively. For each bucket, retrieval considers all candidate pairs between the source and target candidate sets and retains only those satisfying the phonetic similarity criterion defined by \(\Phi\).

Retrieval proceeds through six ordered stages:

\begin{enumerate}
    \item Phonetic matches between original domain terms
    \[
    R_1=\Phi(A,B)
    \]

    \item Phonetic matches between semantic expansions of \(A\) and original \(B\) terms
    \[
    R_2=\Phi(A_S,B)
    \]

    \item Phonetic matches between original \(A\) terms and semantic expansions of \(B\)
    \[
    R_3=\Phi(A,B_S)
    \]

    \item Phonetic matches between semantic expansions of both domains
    \[
    R_4=\Phi(A_S,B_S)
    \]

    \item Phonetic recovery from the \(A\) domain as the sole semantic anchor
    \[
    R_5=\Phi(A_P,A\cup A_S)
    \]

    \item Phonetic recovery from the \(B\) domain as the sole semantic anchor
    \[
    R_6=\Phi(B_P,B\cup B_S)
    \]
\end{enumerate}

Duplicate affordances retain ownership of the earliest stage in which they are discovered. Phonetic retrieval is performed through nearest-neighbor search in the learned IPA embedding space rather than exact homophone lookup, allowing retrieval of approximate sound relationships learned during phonetic embedding training.

\subsubsection{Affordance Ranking and Pruning}

Affordance ranking combines several heuristic quality signals. \textit{Phonetic} scores measure pronunciation similarity, \textit{semantic} scores measure anchorability to the source semantic domains, \textit{naturalness} scores estimate recognizability as French lexical material, and \textit{pivot} scores estimate usefulness for pun construction. Overall scores represent the combined affordance-ranking score used during retrieval pruning.

Several pruning stages are applied before export. Exact duplicates are removed. Trivial inflectional variants, accent variants, plural variants, and same-root morphological echoes are penalized or filtered. Per-surface and per-root caps prevent a single lexical family from dominating the affordance list. Additional constraints limit identity affordances and other low-creativity collisions.

The final retrieval output is a diversified set of higher-scoring retrieved affordances under our filtering criteria, distributed across the expanded retrieval space. These affordances serve as constraints and inspiration for downstream French pun generation. 

\subsection{Candidate Generation}

Our previous system generated a single French pun per input row using the English source text, translated semantic domains, and a small set of retrieved French affordances. However, affordance retrieval covered only 14\% of the dataset, limiting the usefulness of retrieval-guided generation.

The expanded retrieval system increased affordance coverage to 50.8\% of examples and produced between one and six affordances for each successful retrieval. This provided more frequent and diverse guidance to the generation stage.

For \texttt{claude-sonnet-4.6} and \texttt{gemini-3-flash}, the generator received the English pun, translated semantic domains, and retrieved affordances and produced twelve distinct French pun candidates. The prompt encouraged different forms of French wordplay, including homophony, near-homophony, polysemy, idiomatic reinterpretation, phrase-level wordplay, and semantic compensation. Retrieved affordances were presented as creative opportunities rather than required lexical targets, allowing the model to explore multiple ways of exploiting the same underlying semantic and phonetic relationships.

For \texttt{gpt-5.5} and \texttt{gemini-3.1-pro-preview}, a single candidate was generated for each example and passed directly to the ensemble stage. These runs were included to evaluate stronger frontier models with a lower-cost of generation.

This shifted generation from a single-solution paradigm to a generate-and-select paradigm, allowing downstream evaluators to choose among multiple realizations of the same retrieved affordance structure.

\subsection{Two-Stage Multi-Agent Ranking}

The 2025 system evaluated a single generated pun using multiple agents that assigned scores and triggered regeneration when scores fell below a threshold. In contrast, the 2026 system reformulates evaluation as a two-stage ranking problem.

\subsubsection{Stage 1: Intra-Model Candidate Selection}

For \texttt{claude-sonnet-4.6} and \texttt{gemini-3-flash}, the generator produces twelve candidate puns for each source example. Candidate strings are extracted, assigned deterministic five-digit identifiers, stripped of generator metadata, and randomly shuffled before evaluation. The evaluator therefore never observes candidate order or model provenance.

A single judge model is prompted to rank candidates from four independent perspectives \cite{wang2024mmte}\cite{goes2022crowd}:

\begin{itemize}
\item \textbf{Comedian} - Which candidate is funniest to a native French speaker?
\item \textbf{Linguist} - Which candidate exhibits the strongest and most recognizable wordplay?
\item \textbf{Editor} – Which candidate is most natural and fluent in French?
\item \textbf{Translator} – Which candidate best preserves the humorous intent of the English source?
\end{itemize}

Rather than assigning numerical scores, each persona returns an ordered top-five ranking of candidate identifiers. Rankings are then aggregated using weighted Borda voting \cite{saari1995borda}. Candidates omitted from a persona's top-five list receive zero points. The candidate with the highest aggregate Borda score is selected as the Stage~1 winner. The output of Stage~1 is a single winning candidate for each generator model.

\subsubsection{Stage 2: Cross-Model Ensemble Selection}

Next, the Stage~1 winners from \texttt{claude-sonnet-4.6} and gemini-3-flash are combined with single-candidate outputs generated by gpt-5.5 and gemini-3.1-pro. This produces four candidate puns for each source example.

As in Stage~1, candidates are assigned fresh deterministic identifiers, shuffled, and stripped of source-model metadata before evaluation. The same four personas independently rank all four candidates. Persona rankings are again aggregated using weighted Borda voting to produce a final ensemble winner.

The ranking pipeline is executed using \texttt{gpt-5.5}, \texttt{claude-sonnet-4.6}, and \texttt{gemini-3.1-pro} as judge models. We evaluated multiple aggregation strategies:

\begin{enumerate}
\item \textbf{Judges then models} - Aggregate persona rankings within each judge model, then aggregate across judge models.
\item \textbf{Models then judges} - Aggregate judge-model rankings first, then aggregate persona rankings.
\item \textbf{Pooled rankings} - Aggregate all rankings simultaneously within a single Borda computation.
\end{enumerate}

Judge-model weights and persona weights are independently configurable, allowing ensemble variants to emphasize different evaluation criteria. We additionally measure inter-judge agreement, prompt-position exposure effects, and model self-preference using normal, half-weight, and removed-self-vote correction schemes.

\section{Results}

\subsection{Retrieval Statistics}

The retrieval system generated 3,748 affordances for 2,064 of 4,061 source puns, corresponding to 50.8\% coverage (Table~\ref{tab:retrieval_source_summary}). Covered source puns contained an average of 1.82 affordances, while the overall average was 0.92 affordances per source pun. Coverage was slightly higher for homophonic puns (55.1\%) than for homographic puns (50.2\%).

Direct $A_S$--$B_S$ bridges accounted for 2,252 retrieved affordances (60.1\% of the total). Detached phonetic recovery contributed the remaining 1,496 affordances, split between $A_P$-anchored retrieval (719 affordances; 19.2\%) and $B_P$-anchored retrieval (777 affordances; 20.7\%) (Table~\ref{tab:retrieval_source_summary}).

\begin{table}[!htbp]
\centering
\caption{Retrieval output by bucket.}
\label{tab:retrieval_source_summary}
\begin{tabular}{lrrr}
\toprule
Retrieval bucket &
Source puns &
Affordances &
Mean score \\
\midrule
$A_S$--$B_S$ bridge
& 1,287 (31.7\%)
& 2,252 (60.1\%)
& 0.711 \\

$A_P$-anchored retrieval
& 556 (13.7\%)
& 719 (19.2\%)
& 0.626 \\

$B_P$-anchored retrieval
& 587 (14.5\%)
& 777 (20.7\%)
& 0.626 \\

\midrule
Total
& 2,064 (50.8\%)
& 3,748 (100.0\%)
& 0.677 \\
\bottomrule
\end{tabular}
\end{table}

The retrieval inventory was dominated by near-homophonic bridges. Similar-sound relations accounted for 3,344 affordances (89.2\%), while same-sound relations accounted for 404 affordances (10.8\%). Same-sound affordances achieved higher mean overall scores than similar-sound affordances (0.817 vs.\ 0.660). Likewise, bridges discovered between $A_S$ and $B_S$ achieved higher mean overall scores than detached recovery affordances (0.711 vs.\ 0.626).

The affordance count distribution was strongly skewed toward small candidate sets (Table~\ref{tab:affordances_per_pun}). Nearly half of all source puns produced no affordances, while 41.2\% produced one or two affordances. Only 9.6\% of source puns produced three or more affordances.

\begin{table}[!htbp]
\centering
\caption{Distribution of retrieved affordances per source pun.}
\label{tab:affordances_per_pun}
\begin{tabular}{lr}
\toprule
Retrieved affordances & Source puns \\
\midrule
0 & 1,997 (49.2\%) \\
1 &   956 (23.5\%) \\
2 &   717 (17.7\%) \\
3-6 & 391 (9.6\%) \\
\midrule
Total & 4,061 (100.0\%) \\
\bottomrule
\end{tabular}
\end{table}

\subsection{Affordance Utilization}

Tables~\ref{tab:affordance_retention} and \ref{tab:affordance_quality_pipeline} summarize affordance retention and quality across retrieval, generation, and selection stages. Successive selection stages reduced the number of represented affordances while increasing their mean quality scores.

\begin{table}[!htbp]
\centering
\caption{Affordance retention across retrieval, generation, and selection stages.}
\label{tab:affordance_retention}
\begin{tabular}{lrr}
\toprule
Stage &
Puns using affordances &
Distinct affordances used \\
\midrule
Affordance retrieval           & 2,064 (50.8\%) & 3,748 (100.0\%) \\
Stage 1 generation  & 1,935 (47.6\%)          & 3,335 (89.5\%) \\
Stage 1 selection   & 1,167 (28.7\%)          & 1,832 (49.1\%) \\
Stage 2 finalists   &   443 (10.9\%)          &   650 (17.4\%) \\
Stage 2 winners     &   141 (3.5\%)           &   206 (5.5\%) \\
\bottomrule
\end{tabular}
\end{table}

\begin{table}[!htbp]
\centering
\caption{Mean quality scores for affordances used across retrieval, generation, and selection stages. Higher values indicate stronger affordances.}
\label{tab:affordance_quality_pipeline}
\begin{tabular}{lccccc}
\toprule
Stage &
Overall &
Phonetic &
Naturalness &
Semantic &
Pivot \\
\midrule
Affordance retrieval          & 0.677 & 0.846 & 0.520 & 0.892 & 0.520 \\
Stage 1 generation & 0.702 & 0.858 & 0.556 & 0.941 & 0.562 \\
Stage 1 selection  & 0.713 & 0.866 & 0.572 & 0.954 & 0.579 \\
Stage 2 finalists  & 0.716 & 0.866 & 0.578 & 0.968 & 0.585 \\
Stage 2 winners    & 0.724 &
                     0.873 &
                     0.588 &
                     0.978 &
                     0.596 \\
\bottomrule
\end{tabular}
\end{table}

Affordance usage remained dominated by $A_S$--$B_S$ bridges throughout the pipeline. These accounted for 60.1\% of retrieved affordances, 71.8\% and 68.7\% of affordance usage in \texttt{claude-sonnet-4.6} and \texttt{gemini-3-flash} candidates, 73.1\% and 70.4\% after Stage~1 selection, and 72.7\% and 72.5\% among Stage~2 finalists and winners.

In contrast, same-sound affordances were consistently overrepresented relative to their prevalence in the retrieval inventory. Although they comprised only 10.8\% of retrieved affordances, they accounted for 17.6\% and 14.9\% of affordance usage in \texttt{claude-sonnet-4.6} and \texttt{gemini-3-flash} candidates, 22.9\% and 19.4\% after Stage~1 selection, and 23.8\% and 27.5\% among Stage~2 finalists and winners.

Among candidates that used retrieval, \texttt{claude-sonnet-4.6} candidates employed an average of 1.38 affordances and \texttt{gemini-3-flash} candidates employed an average of 1.36 affordances. Ranked candidates showed similar values (1.43 and 1.39, respectively). Most affordance usage involved only one side of a retrieved bridge. In Claude-generated candidates, 48.4\% of matches used only the left affordance, 25.0\% used only the right affordance, and 26.6\% used both bridge endpoints; the corresponding proportions for \texttt{gemini-3-flash} were similar. Candidates that used retrieval employed 1.38 affordances on average for \texttt{claude-sonnet-4.6} and 1.36 for \texttt{gemini-3-flash}, increasing to 1.43 and 1.39 after Stage~1 selection.

\subsection{Candidate Selection Results}

The judge personas exhibited substantial diversity in their preferred candidates. Complete agreement occurred for only 16.8\% of Claude-generated examples and 5.4\% of Gemini-generated examples. Claude examples produced an average of 2.29 distinct persona winners, compared with 2.63 for \texttt{gemini-3-flash}. More than half of \texttt{gemini-3-flash} examples (56.9\%) and 38.7\% of \texttt{claude-sonnet-4.6} examples produced at least three distinct persona winners, indicating that different personas frequently prioritized different candidate qualities. This diversity of preferences motivates the use of rank aggregation rather than relying on a single evaluation perspective.

\subsection{Ensemble Results}

Discriminator models exhibited markedly different selection preferences. The \texttt{gemini-3.1-pro} judge selected Gemini-family candidates 72.4\% of the time, the \texttt{gpt-5.5} judge selected GPT-generated candidates 32.6\% of the time, and the \texttt{claude-sonnet-4.6} judge selected Claude-generated candidates as its top-ranked choice 27.1\% of the time. As shown in Table~\ref{tab:stage2_all_metric_runs}, discriminator choice and persona weighting both substantially affected the final winner distribution, with \texttt{gemini-3.1-pro} judging strongly self-biased and translator-only weighting increasing the share of \texttt{gpt-5.5} winners.

\begin{table}[!htbp]
\centering
\caption{Generator sources selected under different Stage~2 aggregation orders.}
\label{tab:aggregation_order}
\begin{tabular}{lrrrr}
\toprule
Aggregation &
gpt-5.5 &
gemini-3.1-pro &
claude-sonnet-4.6 &
gemini-3-flash \\
\midrule
Judges-first & 1,165 & 1,954 & 507 & 435 \\
Models-first & 1,161 & 1,993 & 463 & 444 \\
Pooled       & 1,176 & 2,066 & 459 & 360 \\
\bottomrule
\end{tabular}
\end{table}

Persona disagreement remained common throughout cross-model selection. Complete persona agreement on the top candidate occurred in only 10.6\% for \texttt{gpt-5.5} judging, 25\% for \texttt{gemini-3.1-pro} judging, and 18.0\% of examples for \texttt{claude-sonnet-4.6} judging. Most examples produced two or three distinct persona winners. Despite disagreement at rank one, personas generally ranked similar candidate sets highly, with all four finalists appearing in every persona's top-four ranking. Pairwise Kendall rank correlations ranged from 0.56 to 0.86 depending on the persona pair \cite{kendall1938tau}.

\begin{table}[!htbp]
\centering
\scriptsize
\caption{Stage~2 winner distributions under selected persona-weighting and discriminator-model configurations. Entries report final winner counts by generator model, with percentages of all 4,061 puns.}
\label{tab:stage2_all_metric_runs}
\begin{tabular}{llrrrr}
\toprule
Persona weights &
Discriminator models &
gpt-5.5 &
gemini-3.1-pro &
claude-sonnet-4.6 &
gemini-3-flash \\
\midrule

$25/25/25/25$ & All (pooled)
& 1,176 (29.0\%) & \textbf{2,066 (50.9\%)} & 459 (11.3\%) & 360 (8.9\%) \\

$45/30/25/10$ & All (pooled)
& 1,003 (24.7\%) & \textbf{1,847 (45.5\%)} & 517 (12.7\%) & 694 (17.1\%) \\

Comedian & All (pooled)
& 944 (23.2\%) & \textbf{1,549 (38.1\%)} & 678 (16.7\%) & 890 (21.9\%) \\

Linguist & All (pooled)
& 1,053 (25.9\%) & \textbf{1,623 (40.0\%)} & 471 (11.6\%) & 914 (22.5\%) \\

Editor & All (pooled)
& 810 (19.9\%) & \textbf{1,886 (46.4\%)} & 626 (15.4\%) & 739 (18.2\%) \\

Translator & All (pooled)
& 1,735 (42.7\%) & \textbf{1,793 (44.2\%)} & 459 (11.3\%) & 74 (1.8\%) \\

\midrule

$25/25/25/25$ & gpt-5.5
& 1,418 (34.9\%) & \textbf{1,630 (40.1\%)} & 483 (11.9\%) & 530 (13.1\%) \\

$45/30/25/10$ & gpt-5.5
& 1,166 (28.7\%) & \textbf{1,441 (35.5\%)} & 541 (13.3\%) & 913 (22.5\%) \\

Comedian & gpt-5.5
& 1,101 (27.1\%) & \textbf{1,310 (32.3\%)} & 656 (16.2\%) & 994 (24.5\%) \\

Linguist & gpt-5.5
& 1,175 (28.9\%) & \textbf{1,404 (34.6\%)} & 461 (11.4\%) & 1,021 (25.1\%) \\

Editor & gpt-5.5
& 888 (21.9\%) & \textbf{1,602 (39.4\%)} & 663 (16.3\%) & 908 (22.4\%) \\

Translator & gpt-5.5
& \textbf{2,126 (52.4\%)} & 1,409 (34.7\%) & 462 (11.4\%) & 64 (1.6\%) \\

\midrule

$25/25/25/25$ & gemini-3.1-pro
& 737 (18.1\%) & \textbf{2,638 (65.0\%)} & 267 (6.6\%) & 419 (10.3\%) \\

$45/30/25/10$ & gemini-3.1-pro
& 637 (15.7\%) & \textbf{2,358 (58.1\%)} & 303 (7.5\%) & 763 (18.8\%) \\

Comedian & gemini-3.1-pro
& 640 (15.8\%) & \textbf{2,121 (52.2\%)} & 409 (10.1\%) & 891 (21.9\%) \\

Linguist & gemini-3.1-pro
& 668 (16.4\%) & \textbf{2,123 (52.3\%)} & 304 (7.5\%) & 966 (23.8\%) \\

Editor & gemini-3.1-pro
& 622 (15.3\%) & \textbf{2,290 (56.4\%)} & 374 (9.2\%) & 775 (19.1\%) \\

Translator & gemini-3.1-pro
& 1,188 (29.3\%) & \textbf{2,476 (61.0\%)} & 300 (7.4\%) & 97 (2.4\%) \\

\midrule

$25/25/25/25$ & claude-sonnet-4.6
& \textbf{1,238 (30.5\%)} & 1,179 (29.0\%) & 1,100 (27.1\%) & 544 (13.4\%) \\

$45/30/25/10$ & claude-sonnet-4.6
& 1,048 (25.8\%) & 1,026 (25.3\%) & \textbf{1,192 (29.4\%)} & 795 (19.6\%) \\

Comedian & claude-sonnet-4.6
& 890 (21.9\%) & 868 (21.4\%) & \textbf{1,364 (33.6\%)} & 939 (23.1\%) \\

Linguist & claude-sonnet-4.6
& \textbf{1,191 (29.3\%)} & 873 (21.5\%) & 1,013 (24.9\%) & 984 (24.2\%) \\

Editor & claude-sonnet-4.6
& 950 (23.4\%) & \textbf{1,301 (32.0\%)} & 1,059 (26.1\%) & 751 (18.5\%) \\

Translator & claude-sonnet-4.6
& \textbf{1,450 (35.7\%)} & 1,391 (34.3\%) & 979 (24.1\%) & 241 (5.9\%) \\

\bottomrule
\end{tabular}
\end{table}

\subsection{Shared Task Results}

At the time of writing, our best submission is ranked first on the public leaderboard with a score of 37.783. Our strongest single-model system is ranked second (37.119), while the best all-model translator ensemble is ranked third (35.949).

The leaderboard ablations show three clear trends. First, GPT-based generation  outperformed the other generator families: \texttt{gpt-5.5} single (37.119) exceeded \texttt{gemini-3.1-pro} single (30.250), \texttt{claude-sonnet-4.6} translator (29.547), and \texttt{gemini-3-flash} translator (23.029). Second, translator-focused selection consistently outperformed other persona weightings. The best translator ensemble achieved 37.783, compared with 25.761 for editor-only, 24.161 for pun-expert-only, and 24.044 for comedian-only selection. Third, candidate pooling was beneficial only when combined with translator-oriented ranking. The all-pooled translator ensemble achieved 35.949, compared with 28.884 for the equal-weight ensemble, 28.493 for judges-first aggregation, and 28.259 for models-first aggregation.

Ensemble weighting also had a large effect. Translator-focused weighting  outperformed comedian- and pun-focused weighting schemes. This result is consistent with our 2025 findings, where systems optimized for semantic fidelity achieved higher BLEU and BERTScore despite often producing weaker target-language wordplay. The official evaluation metric therefore appears to reward preservation of source meaning more strongly than creative pun generation, creating a potential tension between leaderboard optimization and humor quality.

Human evaluation, which is the primary target of our system design, has not yet been performed, and not all participating submissions are visible on the leaderboard. The camera-ready version will report final leaderboard standings and official human evaluation results.

\section{Discussion}

\subsection{Retrieval actually matters}

Retrieved affordances were actively exploited by generators and progressively concentrated among selected winners. As affordance diversity decreased across successive pipeline stages, the mean quality of surviving affordances increased. This suggests that generation and ranking preferentially retained stronger semantic–phonetic collisions rather than treating all retrieved affordances equally.

\subsection{Retrieval is still the bottleneck}

Despite tripling retrieval coverage, nearly half of source puns still yielded no affordances, and retrieved affordances appeared in only 3.5\% of final ensemble winners. This highlights the central challenge of retrieval-guided pun translation: useful sound–meaning collisions are both rare and difficult to discover. Rather than serving as a pervasive generation mechanism, affordances functioned as sparse but valuable opportunities that were exploited only when the retrieval system uncovered particularly strong semantic–phonetic bridges. This difficulty mirrors the challenges faced by human translators, who frequently abandon source-side wordplay when no suitable target-language collision exists.

\subsection{Same-sound affordances are disproportionately successful}

Although exact sound matches constituted only a small fraction of retrieved affordances, they became increasingly overrepresented among selected candidates. Approximate phonetic similarity appears valuable for exploration, but exact phonological collisions are selected at disproportionately high rates when available. This pattern suggests that successful target-language wordplay remains concentrated around rare exact sound correspondences, which provide particularly strong and easily recoverable opportunities for humorous reinterpretation.

\subsection{Affordances cluster around a small number of strong bridges}

Although retrieval produced thousands of affordances, many represented variations of the same underlying semantic–phonetic opportunity. Successful pun translation therefore appears to depend less on the total number of retrieved affordances than on the number of distinct semantic–phonetic bridges available for exploitation. Future work should focus on exploring more distant semantic and phonological neighborhoods capable of surfacing genuinely novel opportunities for wordplay \cite{zhong2024lets}\cite{wang2024innovative}.

\subsection{The persistence of position bias}

Candidate order influenced selection behavior even after randomization, indicating that current LLM judges are not fully position invariant. Reducing these effects will likely require repeated evaluation under multiple candidate orderings, albeit at increased computational cost. The persistence of position effects despite randomization suggests that ranking variance remains a significant source of uncertainty in LLM-based evaluation pipelines.

\subsection{Healthy disagreement is a good thing}

Strong disagreement between judge personas indicates that pun quality is inherently multi-objective. Humor, wordplay strength, fluency, and translation fidelity frequently favored different candidates, suggesting that multi-perspective evaluation captures aspects of quality that would be lost under a single evaluator objective.

Similarly, Gemini-3.1-pro exhibited substantial self-preference, whereas the other judges displayed weaker generator-specific preferences. Multi-judge aggregation therefore reduces reliance on any single discriminator's biases.

\subsection{What the leaderboard is actually rewarding}

Translator-oriented ranking strategies substantially outperformed comedian-, editor-, and pun-focused alternatives on the public leaderboard. This large performance gap suggests that current automatic metrics underweight the qualities that make target-language puns entertaining to human readers and instead place greater emphasis on preservation of source meaning. Notably, the strongest single-model result was obtained by a single \texttt{gpt-5.5} generation rather than a generate-and-select pipeline, suggesting that generator capability remains a major determinant of leaderboard performance. Human evaluation will therefore be necessary to determine whether leaderboard gains correspond to genuine improvements in translated wordplay.

\section{Conclusion}

Our analyses suggest that successful pun translation emerges through a process of discovery, exploration, and selection. Retrieval uncovers semantic–phonetic affordances, generators explore the opportunities they create, and evaluators progressively concentrate around the strongest sound–meaning collisions. 

This picture is remarkably close to the account proposed by Low, who argued that translators need not search for equivalent words, but for new points of contact between meaning and sound. In that sense, the most satisfying outcome of this work is not any particular score or ranking, but the realization of that idea as a computational process. Low's pentagons and hexagons become semantic expansions, phonetic neighborhoods, affordance discovery, and candidate selection. 

The search remains difficult, and many puns still resist translation, but the results suggest that Low's central insight was correct. Successful pun translation is not the preservation of a word. It is the discovery of another place in the language where sound and meaning collide. Languages are full of such hidden bridges. The beauty of pun translation lies in finding them

\section*{Acknowledgements}

We thank the Data Science at Georgia Tech (DS@GT) CLEF competition group for their support.
This research was supported in part through research cyber-infrastructure resources and services provided by the Partnership for an Advanced Computing Environment (PACE) at the Georgia Institute of Technology, Atlanta, Georgia, USA \cite{PACE}. 

\section*{Declaration on Generative AI}
 During the preparation of this work, the authors used \texttt{gpt-5.5} in order to generate LaTeX equations and to check grammar and spelling. After using this tool, the authors reviewed and edited the content as needed and take full responsibility for the publication’s content.

\bibliography{main}

\end{document}